\pdfoutput=1

\documentclass[11pt]{article}

\usepackage[final]{acl}

\usepackage{times}
\usepackage{latexsym}

\usepackage[T1]{fontenc}

\usepackage[utf8]{inputenc}
\usepackage[T2A,LAE,T1]{fontenc}
\usepackage[arabic,greek,USenglish]{babel}
\usepackage{subcaption}

\usepackage{microtype}

\usepackage{inconsolata}

\usepackage{tipa}

\usepackage{graphicx}

\usepackage{todonotes}
\usepackage{booktabs}
\usepackage{ragged2e}
\usepackage{array}
\usepackage{multirow}
\usepackage{amsmath}
\usepackage{tabularx}
\usepackage[shortlabels]{enumitem}

\newcommand{\ARA}[1]{\begin{scriptsize}\AR{#1}\end{scriptsize}}

\title{Revisiting Common Assumptions about Arabic Dialects in NLP}

\author{Amr Keleg, Sharon Goldwater, Walid Magdy\\
Institute for Language, Cognition and Computation \\
School of Informatics, University of Edinburgh \\
\texttt{a.keleg@sms.ed.ac.uk}, \texttt{\{sgwater,wmagdy\}@inf.ed.ac.uk}}

\newcommand\samplesMatching{M}
\newcommand\samplesMatchingValid{M\textsubscript{Val}}
\newcommand\samplesMatchingExc{M\textsubscript{Exc}}
\newcommand\samplesValid{N\textsubscript{Val}}
\newcommand\cuesMatching{C\textsubscript{Mat}}

\begin{document}
    \maketitle

\begin{abstract}
Arabic has diverse dialects, where one dialect can be substantially different from the others.
In the NLP literature, some assumptions about these dialects are widely adopted (e.g., ``Arabic dialects can be grouped into distinguishable regional dialects") and are manifested in different computational tasks such as Arabic Dialect Identification (ADI). However, these assumptions are not quantitatively verified.
We identify four of these assumptions and examine them by extending and analyzing a multi-label dataset, where the validity of each sentence in 11 different country-level dialects is manually assessed by speakers of these dialects.
Our analysis indicates that the four assumptions oversimplify reality, and some of them are not always accurate. This in turn might be hindering further progress in different Arabic NLP tasks.
\end{abstract}

\section{Introduction}
\label{sec:intro}

Arabic has more than 420 million speakers, and is the official language of more than 22 countries, making it the sixth most spoken language worldwide \cite{bergman-diab-2022-towards}. Arabic speakers distinguish between two varieties of the language. Modern Standard Arabic (MSA) is the language of literary work, official documents, and newspapers. MSA has standardized orthography, is taught in schools, and is mostly perceived as a shared variety across Arab countries. Conversely, local dialectal varieties---known as Dialectal Arabic (DA)---are mostly spoken, yet have recently become more written with the rise of social media platforms, despite not having a standardized orthography. These local varieties could differ from MSA and each other in phonology, morphology, syntax, and semantics. 
Different levels are used to group the varieties of DA as varieties spoken into (a)~5-6 macro-regions, (b)~>20 countries, and (c)~>100 cities/provinces.

Variation also exists within the same dialect. To quantify this variation, \citet{keleg-etal-2023-aldi} introduced the Arabic Level of Dialectness (ALDi) metric, defined as how divergent a sentence is from MSA. ALDi is operationalized as a continuous score between 0 (MSA) and 1 (Highly Dialectal), on the level of sentence-like units.

Successful Arabic NLP systems need to handle all of these types of variation, yet some literature rests on certain assumptions about Arabic dialect variation.
In this paper, we identify three common assumptions that were progressively adopted by the Arabic NLP community, in addition to a fourth one that was recently introduced.\footnote{Limitations of the 4 assumptions are discussed qualitatively in the literature but are ignored or perceived as minor.}
The assumptions impact different aspects such as distinguishing between the varieties of DA (\ref{item:validity_region}, \ref{item:sen_length}, and \ref{item:ALDi}), and dialectal samples curation (\ref{item:cues}).
However, their validity is neither backed by enough linguistics studies nor quantitively assessed, making them anecdotal. While they were useful in achieving progress in tasks like Arabic Dialect Identification (ADI)\footnote{As of the 15\textsuperscript{th} of December 2024, 618 papers on Semantic Scholar \cite{10.1038/nature.2015.18703} match ``Dialect Identification", out of which 173 ($\approx$28\%) match ``Arabic Dialect Identification". However, ADI is still unsolved \cite{abdul-mageed-etal-2024-nadi}.}, inaccuracies in these assumptions might hinder further progress.
Our analysis focuses on the text modality, but the findings could apply to the speech modality. It could also benefit linguists studying the Arabic varieties.
We systematically examine the assumptions below:

\begin{enumerate}[label=\textbf{Asm.~\arabic*}, leftmargin=*, itemindent=0pt, wide=\itemindent, topsep=0.5pt, itemsep=-2pt]
    \item \label{item:validity_region} A DA sentence is usually valid in only one regional dialect.

    \item \label{item:sen_length} Only short sentences can be valid in multiple dialects.

    \item \label{item:cues} Distinctive dialectal words (e.g., \ARA{برشة} /brš\textcrh/  for Tunisian Arabic) can be curated to infer the dialect of sentences containing any of them.

    \item \label{item:ALDi} For a sentence valid in multiple dialects, speakers of these dialects consistently provide similar ratings of the sentence's level of dialectness.
\end{enumerate}

In our analysis\footnote{We release our code at: \url{https://github.com/AMR-KELEG/MLADI-assumptions-revisiting}}, we used 978 DA sentences geolocated to 14 different Arab countries. 33 annotators from 11 Arab countries (3 each) labeled each sentence for \textbf{(a)} validity in the annotator's country-level dialects and \textbf{(b)} Arabic Level of Dialectness (ALDi).
We find that >56\% of the dataset is valid in multiple regional dialects, showcasing that ADI is a multi-label classification task (i.e., each sentence should be assigned multiple labels, not a single one). The sentence's ALDi correlates better with the number of its valid dialects than its length. Moreover, lists of dialectal words are not always distinctive of their presumed dialects. Lastly, the ALDi ratings assigned by speakers of different regional dialects can significantly vary, for sentences valid in these dialects. 

\section{Background}
\label{sec:background}
In this section, we describe how the four assumptions were progressively adopted.

\subsection{The groupings of Arabic Dialects}
\label{sec:DA_groupings}
Along the vast geographical area over which Arabic speakers are distributed, different varieties of DA are spoken. Varieties spoken within geographically proximate areas are commonly grouped into regional dialects. An example of such groupings is: the Levant (Lebanon, Jordan, Palestine, Syria), Nile Basin (Egypt, Sudan), Gulf (Saudi Arabia, Oman, Qatar, Bahrain, United Arab Emirates, Iraq), Gulf of Aden (Yemen, Djibouti, Somalia), and Maghreb (Morocco, Tunisia, Algeria, Mauritania, Libya).\footnote{A canonical grouping of the Arabic dialects does not exist \cite{habash2010introduction,abdul-mageed-etal-2018-tweet}.}
Regional groupings recognize the within-region similarities while assuming minimal overlap between the regional varieties.

\paragraph{Regional-level Dialects} Early efforts in ADI used single-label classification to distinguish between a subset of the regional varieties, including MSA as an independent variety/class \cite{biadsy-etal-2009-spoken,zaidan-callison-burch-2011-arabic}.  This adoption of single-label classification implicitly accepts \ref{item:validity_region} at the regional level; i.e., that sentences are usually only valid in one regional dialect.
Three follow-up papers
did back off from this assumption by introducing a new class (\textit{General}) for sentences that are valid in multiple regional dialects
\cite{zbib-etal-2012-machine,cotterell-callison-burch-2014-multi,zaidan-callison-burch-2014-arabic}. The last of these papers found that \textit{General} class represented $\approx 6.3\%$ of the total annotations in their dataset, demonstrating how the regional dialects are not fully distinguishable from each other. However, the authors also noted that
some annotators wrongly selected the \textit{General} class when they could not decide the dialect of the sentence, while others labeled some sentences as only valid in their native dialects although these sentences are valid in other dialects.

Despite these hints of additional complexity,  overlap between the regional dialects was ignored in annotating further datasets \cite{bouamor-etal-2014-multidialectal, salama-etal-2014-youdacc, huang-2015-improved, malmasi-etal-2016-discriminating, zampieri-etal-2017-findings, zampieri-etal-2018-language, el-haj-etal-2018-arabic, alsarsour-etal-2018-dart, abu-farha-etal-2021-overview}.\footnote{See §\ref{sec:regional_level_previous models} for a discussion on regional ADI performance.} A few papers acknowledge this limitation 
of their datasets, providing examples of sentences that are valid in multiple regional dialects \cite{malmasi-etal-2016-discriminating, LULU2018262, Salloum_2018, el-haj-2020-habibi}, or valid in both MSA and a regional dialect\footnote{Some phonological differences are lost in text, making some sentences plausible in both MSA and a variety of DA.} \cite{el-haj-etal-2018-arabic}, but the continued use of single-label annotation implies that these cases are thought to be a small minority.

\paragraph{Country-level Dialects} Grouping dialects into regions abstracts differences between the dialects spoken within each region \cite{9052982,althobaiti2020automaticarabicdialectidentification,10.1007/978-3-031-08277-1_23}, such as those between Egyptian and Sudanese Arabic \cite{abdul-mageed-etal-2018-tweet}, or between the dialects of the Levant \cite{abu-kwaik-etal-2018-shami}. Therefore, more fine-grained sets of labels were proposed for the task of ADI. Country-level ADI is the most common setup \cite{abu-kwaik-etal-2018-shami,9052982, abdul-mageed-etal-2022-nadi,abdul-mageed-etal-2023-nadi},  with some datasets targeting both country-level and province/city-level ADI \cite{abdul-mageed-etal-2018-tweet,salameh-etal-2018-fine,bouamor-etal-2019-madar,abdul-mageed-etal-2020-nadi,abdul-mageed-etal-2021-nadi}.

Country-level ADI has still been modeled as a single-label classification task.
This is problematic as any overlap existing on the regional level will still exist when these regions are divided into countries. Moreover, similar country-level dialects of the same region are expected to overlap. Hence, it has been found that many errors of the country-level ADI models are caused by confusing dialects spoken in neighboring countries, most of which would belong to the same region \cite{biadsy-etal-2009-spoken, salameh-etal-2018-fine, talafha-etal-2019-team, samih-etal-2019-qc, ragab-etal-2019-mawdoo3, priban-taylor-2019-zcu, ghoul-lejeune-2019-michael, eltanbouly-etal-2019-simple, abu-kwaik-saad-2019-arbdialectid, dhaou-lejeune-2020-comparison, talafha-etal-2020-multi, aloraini-etal-2020-qmul, abdelali-etal-2021-qadi, alkhamissi-etal-2021-adapting, el-mekki-etal-2021-bert, jamal-etal-2022-arabic, khered-etal-2022-building, attieh-hassan-2022-arabic}.

\paragraph{Sentence Length and ADI} Most ADI datasets use sentence-like units (e.g., tweets). A common belief (\ref{item:sen_length}) is that most multi-label samples are very short. Since most NLP models would struggle with these short sentences, holding this belief might explain why ADI has continued to be modeled as a single-label classification task.

\subsection{Dialectal Lexical Cues}
\label{sec:extrinstic_factors}

Although dialects differ at many linguistic levels (phonological, lexical, syntactic), one of the easiest types of cues to identify in text is lexical cues \cite{51906}.
These cues are \textit{distinctive} of a particular dialect if they are not shared with other dialects. Some papers provide qualitative examples of these cues like \ARA{هطعش} (/hT\foreignlanguage{greek}{ς}š/ - eleven)\footnote{Transliteration follows HSB scheme \cite{Habash2007}.} for Yemeni \cite{al-shargi-etal-2016-morphologically} and \ARA{برشة} (/brš\textcrh/ - a lot) for Tunisian \cite{McNeil_2018,abdelali-etal-2021-qadi}.

Distinctive cues have been widely used to build DA datasets.
To this end, ad-hoc lists of lexical cues were compiled to collect dialectal samples from websites or social media platforms. These lists were either directly used \cite{al-sabbagh-girju-2012-yadac, alshutayri2017exploring, alshargi-etal-2019-morphologically}, or first validated by speakers of different dialects to ensure their distinctiveness \cite{Almeman2013AutomaticBO, zaghouani-charfi-2018-arap, alsarsour-etal-2018-dart, mubarak2018dial2msa}. 

It is acknowledged that the diversity of the curated samples is limited by the lists of cues \cite{abdul-mageed-etal-2020-toward}. However, the precision and distinctiveness of these cues are assumed to be high without quantitatively measuring them (\ref{item:cues}), which we revisit in this paper.

\subsection{Differences in ALDi Perceptions}

The concept of having different levels of dialectness was noted decades ago \cite{badawi1973mustawayat,10.2307/43192652}. In NLP, two papers designed guidelines for rating the level of dialectness of sentences \cite{habash2008guidelines, zaidan-callison-burch-2011-arabic}. After that, however, the concept was ignored until \citet{keleg-etal-2023-aldi} proposed fine-tuning a BERT-based model to automatically quantify it as a score in [0, 1] on sentence-like units. They found that some sentences can be considered as being closer to MSA or DA based on how an annotator attempts to pronounce them. Therefore, they embrace the variation in the human annotations by averaging them to obtain gold standard ALDi scores. However, this overlooks the impact of the annotator's native dialect on the provided ALDi ratings (\ref{item:ALDi}).

\section{Data}
\label{sec:data}

For our analysis, we release an extended version of the NADI 2024 dataset \cite{abdul-mageed-etal-2024-nadi}, that we call the \textit{\textbf{MLADI}} (Multi-label ADI) dataset.\footnote{We release an accompanying ADI leaderboard at: \url{https://huggingface.co/spaces/AMR-KELEG/MLADI}}
The original NADI 2024 dataset has 1,120 tweets, of which only 70 were automatically identified as MSA and 1,050 as DA. The DA samples' geolocations are uniformly distributed across the 14 most populated Arab countries, excluding Somalia, for which data is not sufficiently abundant. 27 annotators were recruited from 9 Arab countries (3 each): Algeria, Morocco, Tunisia, Egypt, Sudan, Palestine, Syria, Iraq, and Yemen. 
For each sample in the dataset, the annotators \textbf{(a)} identified if a speaker of one of their country-level dialects could have authored the tweet. If an annotator answered (a) as yes, then the sentence is also \textbf{(b)} rated for its ALDi as MSA (L0), Colloquial-influenced MSA (L1),  Normal Colloquial (L2), or Informal (or Vulgar) Colloquial (L3).

\begin{figure}[tb]
    \centering
    \includegraphics[width=\linewidth, trim={1cm 3cm 0.5cm 2.75cm},clip]{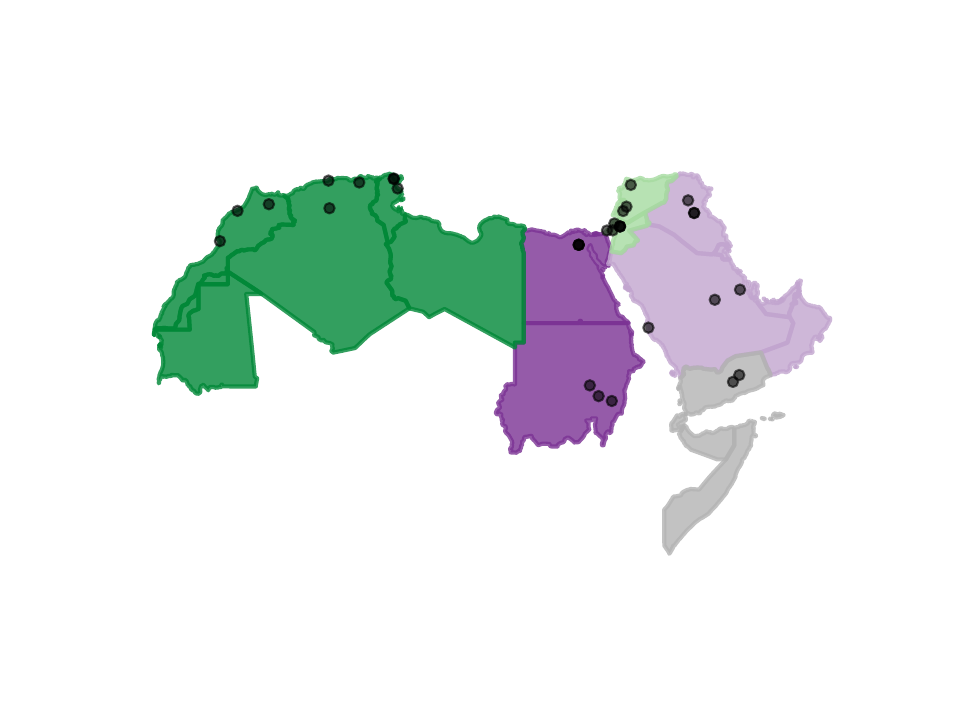}
    \caption{A map of the Arab world. The black dots indicate the provinces/cities from which the annotators originate. 
    Regional dialects (Maghreb, Nile Basin, Levant, Gulf, Gulf of Aden) are encoded as different colors according to the groupings of \newcite{baimukan-etal-2022-hierarchical}.
    }
    \label{fig:arab_world_map}
\end{figure}

The dataset creators provided us with the annotated samples and the individual annotator labels, which we used to study the aforementioned assumptions.
In addition, we recruited 3 annotators from each of Jordan and Saudi Arabia to extend the dataset's labels, using the same annotation guidelines as in \newcite{abdul-mageed-etal-2024-nadi}. This improves the dataset's coverage of the different Arab dialects, especially Gulf Arabic.
\autoref{fig:arab_world_map} shows the annotators' cities/provinces of origin.

The Interannotator Agreement scores (see~§\ref{sec:IAA}) for the two new dialects are similar to the ones reported for the NADI 2024 dataset.
Following the NADI 2024 paper, we use majority voting to identify the validity of each tweet in each of the 11 country-level dialects, and for ALDi, we transform the ratings from discrete levels (L0, L1, L2, L3) into numeric values ($0, \frac{1}{3}, \frac{2}{3}, 1$). A sentence's ratings, for the dialects in which the sentence is valid (according to the majority voting), are averaged to estimate a dialect-agnostic ALDi score.

\section{Analysis}
\label{sec:analysis}
In this section, we investigate each of the four assumptions listed in §\ref{sec:intro}, using 978 out of the 1,050 DA samples, after discarding 72 samples that are not labeled as valid in any of the 11 considered country-level dialects.

\subsection{\ref{item:validity_region} - Arabic Dialects Rarely Overlap}
\label{sec:dialects_overlap}

At least 28 different ADI datasets assign a single regional/country-level dialect to each sentence \cite{keleg-magdy-2023-arabic}. Single-label classification was shown not to be suitable for country-level ADI both qualitatively \cite{kchaou-etal-2019-lium,touileb-2020-ltg,bayrak-issifu-2022-domain,khered-etal-2022-building} and quantitatively \cite{keleg-magdy-2023-arabic, olsen-etal-2023-arabic, abdul-mageed-etal-2024-nadi}. However, single-label classification might still be thought of as suitable for ADI on the level of regional dialects, under the assumption that they rarely overlap.

\paragraph{Method} Using the regional grouping proposed by \newcite{baimukan-etal-2022-hierarchical}, we form 5 regional-level validity labels from the 11 country-level labels as follows: \textbf{1)~Nile Basin (NL):} Egypt, Sudan, \textbf{2)~Gulf (GL):} Iraq, Saudi Arabia, \textbf{3)~Gulf~of~Aden (AD):} Yemen, \textbf{4)~Maghreb (MG):} Tunisia, Algeria, Morocco, and \textbf{5)~Levant (LV):} Jordan, Palestine, Syria. A sentence is valid in a regional dialect if it is valid in at least one of the considered region's countries. Afterward, we count the number of regional dialects in which each sentence is valid.

\begin{figure}[!t]
    \centering
        \includegraphics[trim={0 0.25cm 0 0.25cm},clip]{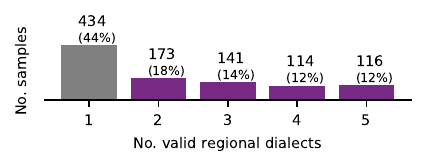}
    \caption{The histogram of the number of valid dialects on the regional level. Only 44\% of the DA samples are confined to single-region dialects.}
    \label{fig:validity_regions}
\end{figure}

\begin{figure}[!t]
    \centering
    \includegraphics[trim={0 0.25cm 0 0.25cm},clip]{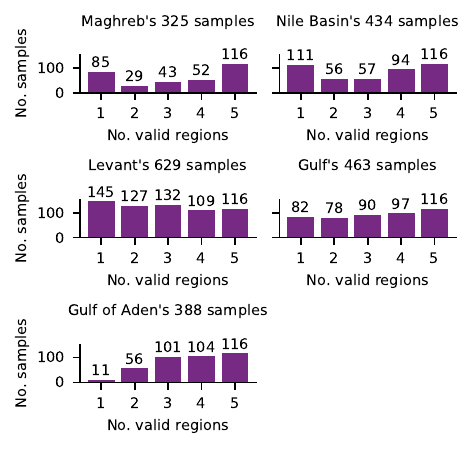}
    \caption{The total number of valid regional dialects for each region's valid samples. \textbf{Note:} The regions' samples are not mutually exclusive (e.g., the same 116 samples valid in the 5 regions are in all distributions).}
    \label{fig:region_valid_dialects_distributions}
\end{figure}

\begin{figure}[!t]
    \centering
    \includegraphics[width=\columnwidth,trim={0.6cm 1.2cm 0 0.25cm},clip]
    {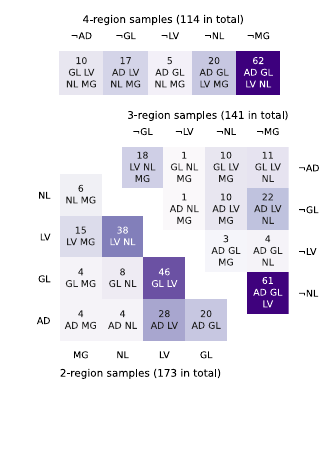}
    \caption{The distribution of the 2-region, 3-region, and 4-region samples across the different combinations. Each combination has its regions indicated in its respective cell. \textbf{Note:} GL/¬GL means valid/not valid in Gulf.}
    \label{fig:regional_samples_distribution}
\end{figure}

\paragraph{Results}A majority 56\% of sentences (544 in total) are valid in multiple regional dialects, as shown in \autoref{fig:validity_regions}. This large cross-regional overlap exists despite the fact the MSA samples were discarded. Notably, 116 of these DA samples (a non-negligible $\sim$~12\%) are valid in all regional dialects.

\paragraph{Further Analysis}
Unlike the other dialects, the Gulf of Aden (represented by Yemen) has only 11 single-region samples as per \autoref{fig:region_valid_dialects_distributions}. Hence, it might not be prominently different from some of the subdialects spoken in other regions, challenging the recognition of \textit{Gulf of Aden} as a regional dialect \cite{habash2010introduction,abdul-mageed-etal-2018-tweet}.

More broadly, \autoref{fig:region_valid_dialects_distributions} shows that the Levant, Gulf, and Gulf of Aden have a substantial number of samples shared with other regional dialects, with Levantine sharing more than the other two dialects.
Looking at the distribution of the multi-region samples in \autoref{fig:regional_samples_distribution}, a large number of the 2-region samples are between pairs of these three regions (e.g., 46 valid in GL and LV, 20 valid in AD and GL) and a majority of 61 samples of the 3-region ones are valid in these regions.
Additionally, LV has a substantial number of 38 samples shared with NL, 15 shared with MG, and 18 shared with both. This explains how LV shares more samples with other dialects than GF and AD.

For the remaining two dialects, both share fewer samples with other dialects, with NL sharing more samples than MG. 62 samples (a majority of the 4-regions samples) are valid in all regions but MG. This is a sign of the dichotomy between the Eastern dialects of Arabic spoken in the Maghreb and the other dialects spoken in the West of the Arab world \cite{51906}. Still, MG shares more with other dialects than previously assumed.

\paragraph{Implications}
Substantial overlap exists between the regional dialects, which contradicts the general perception that they are distinguishable from each other. As previously mentioned, this overlap will still exist when the regions are split into countries as shown in~§\ref{sec:country_level_overlap}. Hence, ADI is a multi-label task on both the regional and country levels.

Classifying \textit{Gulf of Aden} as a distinct regional variety requires reevaluation, given the limited number of samples only valid in this region. Similarly, dialectal categorizations that are not based on the country borders could be considered.\footnote{\href{https://glottolog.org/resource/languoid/id/arab1395}{Glottolog} and \href{https://www.ethnologue.com/language/ara/}{Ethnologue} recognize 37 and 28 Arabic dialects,  respectively.}

\subsection{\ref{item:sen_length} - Only Short Sentences' Dialects are Ambiguous}
\label{sec:sentence_length}

\begin{figure*}[!t]
    \begin{subfigure}{\textwidth}
        \includegraphics[]{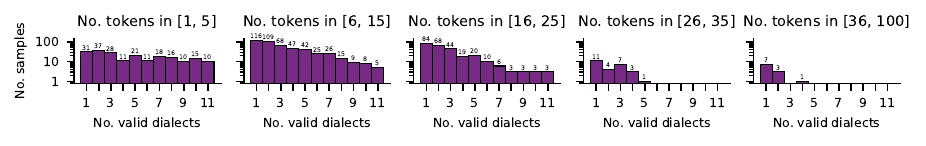}
        \caption{Sentence length (measured as the number of tokens). \textbf{Note:} $\rho(Sentence\ Length,No.\ valid\ dialects) = -0.28$}
        \label{fig:validity_length}
    \end{subfigure}
    
    \begin{subfigure}{\textwidth}
        \includegraphics[]{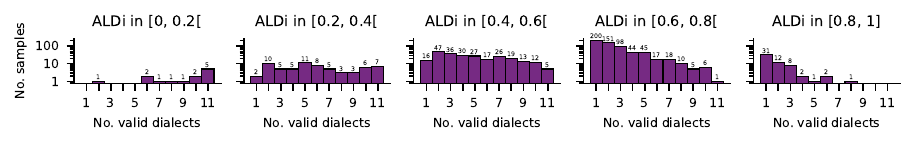}
        \caption{ALDi scores (averaged across all ratings).
        \textbf{Note:} $\rho(ALDi,No.\ valid\ dialects) = -0.52$}
        \label{fig:validity_ALDi}
    \end{subfigure}
    \caption{The distribution of the sentences (log scale) and the number of valid country-level dialects according to different ranges of sentence length \textbf{(a)} and ALDi scores \textbf{(b)}. \textbf{Note:} Since the MSA samples were automatically discarded from our analysis dataset, there are very few samples with low ALDi scores ($\in [0, 0.2]$). However, the histogram of this bin is expected to be left-skewed (i.e., MSA samples are expected to be valid in all dialects).}
    \label{fig:validity_factors}
\end{figure*}

In the context of ADI, sentence length is discussed from two points of view (POVs). \textit{POV \#1} explicitly mentions that the dialect of extremely short speech segments/text sentences can be ambiguous.
Hence, it is infeasible for humans, and consequently machines, to assign a single dialect to these segments \cite{alorifi_2008} and sentences \cite{el-haj-etal-2018-arabic, alsarsour-etal-2018-dart, abu-kwaik-saad-2019-arbdialectid, althobaiti_2022}.
\textit{POV \#2} empirically finds that the
longer the segment/sentence gets, the higher the performance of a single-label ADI system is, for speech \cite{biadsy-etal-2009-spoken, 9052982} and text \cite{zaidan-callison-burch-2014-arabic,salameh-etal-2018-fine, alkhamissi-etal-2021-adapting, abdelali-etal-2021-qadi, bayrak-issifu-2022-domain}.
This can be attributed to a decline in dialect ambiguity as sentences get longer.

\paragraph{Method} We examine the assumption by computing Spearman's correlation between the sentence length (as the number of tokens) and the number of valid dialects on the country level. Additionally, we study the histograms of the number of valid dialects for five different ranges of sentence lengths.

\paragraph{Results}
According to \autoref{fig:validity_length}, the majority of trivially short sentences are valid in multiple dialects as per \textit{POV \#1}. However, \textit{POV \#1} overlooks the large number of moderately long sentences (16-25 tokens) that are also valid in multiple dialects. Additionally, despite long sentences being valid in a smaller number of dialects, 
confirming \textit{POV \#2}, there is only a weak negative Spearman's correlation coefficient (-0.28) between the sentence length and its number of valid dialects.

\paragraph{Further Analysis} On replicating the analysis by replacing the sentence length with the ALDi score, a stronger negative correlation (-0.52) is realized.\footnote{
A coefficient of -0.45 is realized when replacing the aggregated manually-assigned ALDi scores with ones automatically estimated using the \href{https://huggingface.co/AMR-KELEG/Sentence-ALDi}{Sentence-ALDi model} \cite{keleg-etal-2023-aldi}.
}
\autoref{fig:validity_ALDi} also indicates that sentences of ALDi scores < 0.2 are generally valid in most of the dialects. Samples with ALDi scores $\in [0.2, 0.4[$ seem to be evenly probable across the different number of validity labels. The distribution then shifts to be more and more right-skewed for the subsequent ranges of ALDi scores.

\paragraph{Implications} Previous assumptions about sentence length are either incomplete (\textit{POV \#1}) or not sufficiently accurate (\textit{POV \#2}). Moreover, a sentence's ALDi score correlates moderately with the number of dialects in which it is valid, making it a better predictor than sentence length. As a proxy of a sentence's number of valid dialects, ALDi could guide the predictions of a multi-label ADI system.

\subsection{\ref{item:cues} - Dialects' Distinctive Lexical Cues}
\label{sec:cues_analysis}

\begin{table*}[!t]
    \begin{subtable}[t]{0.95\columnwidth}
    \centering
    \footnotesize
    \addtolength{\tabcolsep}{-2.5pt}
    \begin{tabular}{c|cccc|ccc|cc}

        \textbf{Region} & \textbf{\samplesMatching} & \textbf{\samplesMatchingValid} & \textbf{\samplesMatchingExc} & \textbf{\samplesValid} & \textbf{P} &  \textbf{D} & \textbf{R} & \textbf{C} & \textbf{\cuesMatching} \\
        \midrule
EGY  &       60 &                  36 &                21 &               287 & .60 & .35 & .13 & 271 & 28\\
IRQ &        7 &                   6 &                 6 &               204 & .86 & .86 & .03 & 120 & 7 \\
        \midrule
MGH &       21 &                  16 &                14 &               325 & .76 & .67 & .05 & 273 & 13\\
LEV &       32 &                  29 &                25 &               629 & .91 & .78 & .05 & 240 & 11\\
 GLF &        9 &                   0 &                 0 &               407 & .00 & .00 & .00 & 200 & 3 \\
         \bottomrule
    \end{tabular}
    \addtolength{\tabcolsep}{+2.5pt}    

        \label{tab:precision_DART}
        \caption{DART's 5 regional lists.}
    \end{subtable}%
    \hspace{0.7cm}
    \begin{subtable}[t]{0.95\columnwidth}
    \centering
    \footnotesize
    \addtolength{\tabcolsep}{-2.5pt}    
    \begin{tabular}{c|cccc|ccc|cc}
        \textbf{Region} & \textbf{\samplesMatching} & \textbf{\samplesMatchingValid} & \textbf{\samplesMatchingExc} & \textbf{\samplesValid} & \textbf{P} &  \textbf{D} & \textbf{R} & \textbf{C} & \textbf{\cuesMatching} \\
         \midrule
		EGY & 53 &                  43 &                20 &               287 & .81 & .38 & .15 & 28& 19\\
        \midrule
		MGH &       45 &                  36 &                31 &               325 & .80 & .69 & .11 & 60 & 26\\
		LEV  &       38 &                  34 &                34 &               629 & .89 & .89 & .05 & 31 & 11\\
 		 GLF &         0 &                   - &                 - &               407 &         - &            - & .00 & 9 & 0 \\
		 \bottomrule
    \end{tabular}
    \addtolength{\tabcolsep}{+2.5pt}    
        \label{tab:precision_DIAL2MSA}
        \caption{DIAL2MSA's 4 regional lists.}
    \end{subtable}%

    \caption{The Precision (\textit{P}), Distinctiveness (\textit{D}), and Recall (\textit{R}) of each region's cues. \textbf{Note:} For each region's list, we report the number of samples of our dataset matching any of the cues (\textit{\samplesMatching}) of which valid (\textit{\samplesMatchingValid})  and of which exclusively valid (\textit{\samplesMatchingExc}), in addition to the total number of valid samples (\textit{\samplesValid}). The last two columns represent the total number of regional cues (\textbf{C}) and the number of cues that match any of the samples (\textbf{\cuesMatching}).
}
    \label{tab:precision_cues}
\end{table*}

\paragraph{Method} For each of DART's \cite{alsarsour-etal-2018-dart} and DIAL2MSA's \cite{mubarak2018dial2msa} lists of regional-level distinctive cues, we identify sentences of our dataset that match at least one of the lexical cues.\footnote{We could not get access to the lists of \cite{Almeman2013AutomaticBO, zaghouani-charfi-2018-arap, alshargi-etal-2019-morphologically}.} We normalize the sentences and lists of cues to handle common typos/ dialectal variations of the same characters (e.g., \ARA{ة} is normalized to \ARA{ه} and \ARA{أ}, \ARA{آ}, \ARA{إ} are normalized to \ARA{ا}) \cite{03463cbd-280f-3a72-b446-e2ff8514989b,INR-031}. Exact matching is then used between the lexical cues and the whitespace tokenized sentences' tokens.
 
For each dialect, we report the number of samples matching at least one of its distinctive cues (\textit{\samplesMatching}). Then, we count the number of matching samples manually annotated as valid in this dialect (\textit{\samplesMatchingValid}), and the number of matching samples that are only (i.e., exclusively) valid in this dialect (\textit{\samplesMatchingExc}). Precision~(P), Distinctiveness~(D), and Recall~(R) of each list are computed as $P=\frac{\samplesMatchingValid}{\samplesMatching}$, $D=\frac{\samplesMatchingExc}{\samplesMatching}$, and $R=\frac{\samplesMatchingValid}{\samplesValid}$; where (\textit{\samplesValid}) is the total number of samples valid in the considered dialect.

Adhering to the regional groupings used in both lists, we aggregate the 11 country-level validity labels into the following regions: \textbf{1)~Egypt}, \textbf{2)~Iraq}, \textbf{3)~Gulf:} Saudi Arabia, \textbf{4)~Maghreb:} Algeria, Morocco, Tunisia, \textbf{5)~Levant:} Jordan, Palestine, Syria. The dialects of Sudan and Yemen were ignored in both lists, so we considered them as \textbf{6)~Others}.

\paragraph{Results} \autoref{tab:precision_cues} shows the results. The extremely low range of recall values for both manually validated lists confirms that relying on these lists of cues limits the number of matching samples. Conversely, the range of the precision scores is generally high (yet not perfect),
except for the cues of Gulf Arabic.
The Egyptian Arabic cues have a low precision score (0.6) for DART and extremely low distinctiveness values (0.35 and 0.38) for both lists.

The samples' validity in the Maghreb, Levant, and Gulf regions is only defined by the subset of the region's countries from which we could recruit annotators. Hence, the precision scores for these regions might improve after collecting annotations for more country-level dialects. However, the non-perfect Distinctiveness scores indicate that some cues of these regions are used in other regional dialects, even when the cues were manually validated for their distinctiveness by the lists' creators.

\paragraph{Qualitative Analysis} On manually inspecting the matching samples, we found that DART's three matching cues of Gulf Arabic (\ARA{شنو}~/šnw/, \ARA{علامك}~/\foreignlanguage{greek}{ς}lAmk/, \ARA{مواعين}~/mwA\foreignlanguage{greek}{ς}yn/) are indeed dialectal terms that are valid in other regional dialects, hence are not indicative of Gulf Arabic.
Additionally, other terms are false friends, having different meanings in MSA and DA varieties, and are not distinctive of a specific dialect in the absence of context.
For instance, the terms (\ARA{ماشي}~/mAšy/ and \ARA{حد}~/Hd/) have the meanings \textit{okay} and \textit{someone} in Egyptian Arabic. However, they have different meanings in MSA (\textit{walking} and \textit{limit}).  The MSA sense of these terms could be used in the context of other dialects, as demonstrated in 
the examples below, which both use the term \ARA{حد}~/Hd/ (underlined in the examples). Example (1) uses this term with its Egyptian meaning ({\it someone}) and is labeled as valid in Egyptian, whereas (2) uses the term with its MSA meaning ({\it limit}) and is labeled as valid in Algerian and Tunisian. Therefore, the term \ARA{حد}~/Hd/ cannot be considered a valid cue to Egyptian Arabic, as assumed in DART.
 
 \vspace{3mm}
\hspace{-1.7\parindent}
\begin{tabularx}{\columnwidth}{l @{} X}   
    (1) &    \multicolumn{1}{r}{.\ARA{ اهله عرفوا يربوه وحد تاني منفعش فيه التربيه} \underline{\ARA{حد}} \ARA{دا الفرق بين}}\\
& `This is the difference between a well-mannered and a bad-mannered person.' \\
    & \\
    (2) &     \multicolumn{1}{r}{.\ARA{الان مازال حظوظ تونس كبيره هاك تراو الفرق الكبار} \underline{\ARA{حد}} \ARA{الي} } \\
     & `So far, Tunisia still has great chances, this is how big teams are.'\\
    \end{tabularx}

\paragraph{Implications} More rigor is needed in building lists of distinctive dialectal words, especially when the curated sentences need to be surely valid in a specific dialect and/or exclusively valid in this dialect.
Using a second validation step (e.g., information about the geolocation of the sentence's author) could increase the precision of the dialects assigned based on the cues' associated dialects. However, this does not ensure distinctiveness and further decreases the recall (see~§\ref{sec:lexical_cues_geolocation}).

\subsection{\ref{item:ALDi} - ALDi Perceptions across Dialects}

Inspired by earlier work \cite{zaidan-callison-burch-2011-arabic}, \newcite{keleg-etal-2023-aldi} introduced the idea of ALDi prediction as an important task. Two recent datasets provide pairs of sentences with their corresponding aggregated ALDi scores: \textit{AOC-ALDi} \cite{keleg-etal-2023-aldi} and \textit{NADI~2024} \cite{abdul-mageed-etal-2024-nadi}. For the former, three annotations per sentence were sought by randomly assigning the sentences to speakers of different dialects \cite{zaidan-callison-burch-2011-arabic}. For the latter, 27 annotators rated the ALDi of each sentence only when it was valid in their country-level dialect. Both datasets used the mean of a sentence's ALDi ratings as its gold-standard ALDi score.
The implicit assumption is that ALDi scores do not depend on the annotator's native dialect; however, this has not been empirically validated. We have shown (\S\ref{sec:sentence_length}) that even sentences with moderate ALDi scores can be valid in multiple dialects, but it is possible that the scores assigned by annotators from those dialects could systematically differ.

\paragraph{Method} 
We compute the Mean Difference (MD) of country-level ALDi scores for each pair of countries.
MD is computed for a pair of countries $r$ and $c$, with $N_{rc}$ sentences valid in both, as
\[\text{MD}(r,c)=\frac{1}{N_{rc}}\sum_{i=1}^{N_{rc}}{(\text{ALDi}_{r}[i] - \text{ALDi}_{c}[i]),}\] where $\text{ALDi}_{r}[i]$ and $\text{ALDi}_{c}[i]$
are the averages of sentence \textit{i}'s ALDi ratings provided by the annotators of \textit{r} and \textit{c} respectively.

\begin{figure*}[!tb]
    \centering
    \includegraphics[width=1.5\columnwidth,trim={0cm 0.3cm 0 0.25cm},clip]{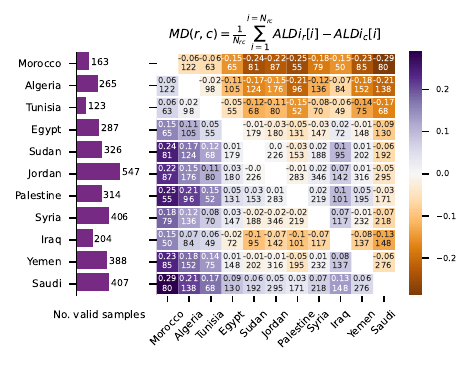}
    \caption{\textbf{(Left)}~The number of valid samples per country (with countries ordered such that same-region ones are consecutive). \textbf{(Right)}~Mean difference (MD) of row country's (\textit{r}) and column country's (\textit{c}) ALDi scores, for the $N_{rc}$ sentences valid in both ($N_{rc}$ is shown as the bottom number in each cell).}
    \label{fig:MD_countries}
\end{figure*}

\paragraph{Results}
\autoref{fig:MD_countries} summarizes the results.
The top three (orangish) rows
indicate that when sentences are valid in one of the Maghreb's countries and another non-Maghrebi country, the annotators from the Maghrebi country rate these sentences to be less dialectal than the non-Maghrebi ones. The difference (e.g., MD(Morocco, Saudi)~=~-0.29) can be close to $\frac{1}{3}$, which is the difference between two consecutive levels of ALDi ratings ($0, \frac{1}{3}, \frac{2}{3}, 1$).
A similar pattern holds true for Iraq to a lesser extent. 
Conversely, Saudi annotators assign higher ALDi scores to sentences common with other dialects.
Many of the country-level differences are statistically significant, with Standard Errors < 0.035. 
However, these differences could arise simply because the annotators differ randomly in their mean scores, independent of dialect. So we might see an apparent difference between country groups if we happened to get annotators with higher means in some countries than in other countries. Due to having only three
annotators per country, 
it is not possible to conclusively test for an effect of dialect (separate from annotator) at the country level, although the consistent trends in the visualization are suggestive. Instead, we test for regional-level differences between annotators, as described below. If additional annotations from each country are obtained in the future, a similar test
could be used at the country level.

\paragraph{Statistical Analysis}
We use a one-sided permutation test to assess whether the differences between two groups of annotators (G\textsubscript{A}, G\textsubscript{B}), of sizes |G\textsubscript{A}| and |G\textsubscript{B}| respectively, can be attributed to the groups' dialects.
First, we compute the MD score between the observed groups' mean ALDi scores (MD\textsubscript{obs}), for the N\textsubscript{AB} sentences valid in both groups.
A large number of pairs of groups \{(A$^\prime$,~B$^\prime$)\} with sizes |G\textsubscript{A}|, |G\textsubscript{B}| are sampled (50k in our case). The pairs of groups (A$^\prime$,~B$^\prime$) are formed by random shuffling and distributing all the annotators across two groups. MD scores for each pair are computed for the same N\textsubscript{AB} sentences.\footnote{In some permutations, we discard the small proportion of sentences that have no ALDi ratings for one of the groups.}
The $p$-value is the percentage of the shufflings with MDs~$\le$~the~observed~grouping's~mean~difference~(MD\textsubscript{obs}).

We consider the annotators of each region as a group, merging Gulf and Gulf of Aden into one region based on the findings of §\ref{sec:dialects_overlap}. Accordingly, we find significant MDs of -0.09, -0.13, -0.14 between the ALDi scores averaged across the annotators of Maghreb against those of Nile Basin, Levant, and Gulf/Gulf of Aden, with p-values of 0.007, 0.00002, and 0.0002, respectively. Similarly, Nile Basin's annotators provide significantly lower ALDi scores than Levantine annotators, with MD of -0.05 (p-value~=~0.04). Differences between other pairs are not statistically significant.

\paragraph{Discussion} There is a general impression that the Arabic dialects are not equally distant from MSA, with some researchers claiming certain dialects---e.g., Gulf Arabic \cite{zaidan-callison-burch-2014-arabic} and Palestinian Arabic \cite{KWAIK20182}---are closer to MSA than others, which could explain the MDs we found for samples shared between different countries/regions. 

\paragraph{Implications} 
Further analysis is required before taking these MDs as an objective measure of a variety's divergence from MSA. Figure~\ref{fig:region_valid_dialects_distributions} indicates that all regions---except \textit{Gulf of Aden}---have many samples not shared with other regions. Single-region samples could still be highly divergent from MSA.
Moreover, people's perception of dialectness is influenced by how they use MSA terms colloquially. For example, both \ARA{خمر} /xmr/ and \ARA{خمرة} /xmr\textcrh/ are valid MSA terms for \textit{wine}. The Holy Qur'an mentions the former, while the latter is more colloquially used in Egypt. Hence, Egyptians might link the first to CA/MSA, and the latter to DA. Consider some MSA lexical items that are shared with dialect \textit{D\textsubscript{A}} but not with dialect \textit{D\textsubscript{B}}. Sentences with these items could be rated as more dialectal by speakers of \textit{D\textsubscript{A}} than \textit{D\textsubscript{B}}. Lastly, sentences valid in multiple varieties could share the same surface form but have different semantics in each variety.

\section{Further Implications in NLP}

Recent improvements to how the varieties of Arabic are computationally modeled \citep{keleg-etal-2023-aldi,keleg-magdy-2023-arabic,abdul-mageed-etal-2024-nadi} are being used in multiple applications, such as better routing of samples to annotators \citep{keleg-etal-2024-estimating}, evaluating the LLMs' dialectal capabilities \citep{robinson2025alqasidaanalyzingllmquality}, and building better recommendation systems \citep{alshabanah-annavaram-2025-using}. Hence, validating widely-held assumptions about Arabic could lead to further progress in automatic ADI and many other tasks/applications.

For example, Arabic NLP researchers used manually curated lists of words/phrases to curate data for various applications like compiling dialect-specific pretraining data \citep{Gaanoun2024}, creating datasets for sentiment analysis \citep{refaee-rieser-2014-arabic}, and offensive text classification \citep{chowdhury-etal-2020-multi}. Therefore, our finding---that some terms share the same orthographic form but have different semantic meanings/senses in various varieties of Arabic---has implications for building datasets for tasks beyond ADI.

Moreover, parallels of the first three assumptions exist beyond Arabic. For example, the overlap between different dialects of the same language has already been noted for other languages such as English, French, and Spanish \citep{bernier-colborne-etal-2023-dialect,zampieri-etal-2024-language,lopetegui-etal-2025-common}. Our findings argue for modeling
dialect identification as a multi-label classification task, even on macro-regional levels.
In addition, sentence length has been discussed as an important predictor of language identification models' performance \cite{baldwin-lui-2010-language}, especially for closely-related languages and dialects \citep{tiedemann-ljubesic-2012-efficient,blodgett2017racialdisparitynaturallanguage,kanjirangat-etal-2022-early}.
We show that the conscious \textit{Dialect Level} choice that Arabic speakers make---operationalized as ALDi---is a better predictor of the number of dialects in which a sentence is valid than its length. Speakers of other languages make similar conscious decisions about how much they adhere or diverge from the standard variety of their language (e.g., \citealp{shoemark-etal-2017-aye}). For these languages, modeling the sentences' divergence from the language's standard variety, as ordinal/quantitative variables, could also provide better predictors of a sentence's validity in multiple dialects than the sentence's length.

\section{Conclusion and Moving Forward}
We identified four common assumptions regarding
Arabic dialects, and systematically studied them by extending
the annotations of a previous dataset to cover more country-level dialects.
Our analysis shows that these assumptions oversimplify some details that, in turn, impact how tasks are framed, datasets are created, and models are trained.

In particular, our main findings and recommendations are as follows.
(1) Arabic dialects overlap considerably at both the country and regional levels, so ADI should be modeled as a multi-label task at both levels. (2) Existing lists of supposedly distinctive lexical cues are less distinctive than previously thought. More rigorous validation is needed for such lists in the future. (3) ALDi scores (but not sentence length) provide a good proxy of a sentence's validity in multiple dialects, which could be used to inform annotation and modeling decisions. Nevertheless, researchers should be aware that speakers of different dialects may systematically differ in their ALDi annotations of the same sentences. (4) Future work should study if sentences with diverging ratings by speakers of different dialects have different semantic meanings in these dialects.

\section*{Limitations}
This paper revisits some widely-held and mostly unquantified assumptions about the Arabic dialects by extending the annotations of the NADI 2024 dataset to have better coverage of the dialects.
Replicating the analysis on other datasets would provide more evidence for the generalizability of our results. Moreover, extending our analysis to cover more country-level dialects might uncover more results than the ones we had when considering 11 country-level dialects. The same applies to using a more granular grouping of the Arabic dialects like different dialects spoken within the same country (e.g., city-level/province-level dialects).

Despite having three annotators per country, our crowdsourced annotators are skewed toward younger age groups and have/are pursuing higher education degrees. Therefore, we acknowledge that our results could be representative of the perceptions of specific demographics within each country.

The analyzed tweets' geolocations are uniformly balanced across 14 different Arab countries, covering a wide range of Arabic dialects. However, we acknowledge that some sub-dialects are not well represented online, as shown by \newcite{mohamed-eida-etal-2024-well} for the Sa'idi Arabic variety of Egypt. Moreover, the data does not have Arabic sentences written in Latin script (known as Arabizi). Arabizi is prominently used in the Maghreb region \cite{10.1007/978-3-319-24800-4_1}, and to a lesser extent in other countries such as Lebanon and Egypt \cite{tobaili-2016-arabizi}.

\section*{Acknowledgments}
We thank Adam Lopez, Nina Gregorio, Burin Naowarat, Yen Meng, Oli Liu, and Sung-Lin Yeh for their comments on an earlier draft of this paper. This work was supported by the UKRI Centre for Doctoral Training in Natural Language Processing, funded by the UKRI (grant EP/S022481/1) and the University of Edinburgh, School of Informatics.

\section*{Ethical Considerations}
The NADI 2024 dataset, which we extended and used in our analysis, has a few samples with offensive language. Our annotators were asked to provide consent confirming their agreement to annotate these samples at the start of the annotation process. The annotation process we followed was approved by the Research Ethics Committee of the University of Edinburgh, School of Informatics, with reference number 839548.

\bibliography{anthology,custom,custom_ar}

\onecolumn
\appendix
\setcounter{table}{0}
\setcounter{figure}{0}
\renewcommand{\thetable}{\Alph{section}\arabic{table}}
\renewcommand{\thefigure}{\Alph{section}\arabic{figure}}

\section{Was Regional-level ADI Already Solved?}
\label{sec:regional_level_previous models}

When framing a multi-label task as a single-label one, there is an expected maximal accuracy that an oracle model can achieve. For a sample with multiple valid labels, the gold-standard label and the prediction of the oracle model will both be randomly selected from the sample's set of valid labels. Both the randomly sampled gold standard label and the model's prediction should match for the prediction of the model to be considered correct. \citet{keleg-magdy-2023-arabic} introduced \autoref{eq:generic} for estimating the expected maximal accuracy given the distribution of the number of labels in which a sentence is valid. Applying the formula to the regional-level labels of the 978 DA samples we used for our analysis, we get an expected maximal accuracy of 63.06\% as per \autoref{eq:NADI2024_maximal_accuracy}. Such a low accuracy upper bound provides more evidence for modeling the task as a multi-label classification one.
\begin{equation}
\label{eq:generic}
    \small
        \mathbf{E[Accuracy_{max} (Dataset)] =\ }
        Perc_{1} + \sum_{n=2}^{n=N_{dialects}}{\frac{Perc_{n}}{n}} 
\end{equation}

\begin{equation}
\label{eq:NADI2024_maximal_accuracy}
    \small
        \mathbf{E[Accuracy_{max} (NADI\ 2024_{regional})] =\ }
        44 + \frac{18}{2} + \frac{14}{3} + \frac{12}{4} + \frac{12}{5} \approx 63.06\%
\end{equation}

\begin{table}[!h]
    \centering
    \scriptsize
    \begin{tabular}{ll}
         \textbf{Test Set(s) Information and Label Distribution} & \textbf{Results}\\
         \midrule
         \textbf{- AOC \textsuperscript{\textbf{*}}}: A random 10\% of the dataset (>110K samples) & \multirow{2}{*}{Acc = 81\%\textsuperscript{$\dagger$}} \\
         \textit{MSA} (>60\% of the samples) - \textit{EGY} - \textit{LEV} - \textit{GLF} & \\
         \cite{zaidan-callison-burch-2014-arabic} & \\
         \midrule
         \textbf{- AOC}: \textit{MSA} (6,355) - \textit{EGY} (1,050) - \textit{LEV} (1,050) - \textit{GLF} (1,050)&  Acc = 87.8\%\textsuperscript{$\dagger$}\\
         \textbf{- FB test set}: \textit{MSA} (1,363) - \textit{EGY} (800) - \textit{LEV} (123) - \textit{GLF} (96) & Acc = 68.2\%\\
         \cite{huang-2015-improved} & \\
         \midrule
         \textbf{VarDial 2016}: \textit{MSA} (274) - \textit{EGY} (315) - \textit{LEV} (344) - \textit{GLF} (256) - \textit{NOR} (351) & Acc = 51.2\%\textsuperscript{$\dagger$}\\
         \cite{malmasi-etal-2016-discriminating}&\\
         \midrule
         \textbf{VarDial 2017}: \textit{MSA} (262) - \textit{EGY} (302) - \textit{LEV} (334) - \textit{GLF} (250) - \textit{NOR} (344)  & F1\textsubscript{weighted} = 0.763 \textsuperscript{\textbf{Sp}} \\
         \cite{zampieri-etal-2017-findings} & \\
         \midrule
         \textbf{VarDial 2018 (Broadcast)}: \textit{MSA} (262) - \textit{EGY} (302) - \textit{LEV} (334) - \textit{GLF} (250) - \textit{NOR} (344) & \multirow{2}{*}{F1\textsubscript{macro} = 0.589 \textsuperscript{\textbf{Sp}}} \\
         + \textbf{VarDial 2018 (YouTube)}:  \textit{MSA} (944) - \textit{EGY} (1,143) - \textit{LEV} (1,131) - \textit{GLF} (1,147) - \textit{NOR} (980) \\
         \cite{zampieri-etal-2018-language} & \\
         \midrule
         
        \textbf{MADAR (CORPUS-6)}: \textit{MSA} (2,000) - \textit{BEIRUT} (2,000) - \textit{CAIRO} (2,000) - \textit{DOHA} (2,000) - \textit{TUNIS} (2,000) - \textit{RABAT} (2,000)& \multirow{2}{*}{Acc. = 93.6\%\textsuperscript{$\dagger$}} \\
        \cite{salameh-etal-2018-fine} & \\
         \midrule

        \textbf{Arabic Dialects Dataset}: A subset of AOC and a Tunisian Corpus & \multirow{2}{*}{Acc = 66.12\%\textsuperscript{$\dagger$}}\\
        \textit{EGY} (1,741) - \textit{GLF} (1,092) - \textit{LEV} (1,056) - \textit{MSA} (1,600) - \textit{NOR} (1,584) & \\
        \cite{el-haj-etal-2018-arabic} & \\
         \midrule

         \textbf{Habibi \textsuperscript{\textbf{*}}}: A random 30\% of the Habibi dataset (50,550 samples)& \multirow{3}{*}{Acc = 72.6\%\textsuperscript{$\dagger$}}\\
         Egyptian (27.7\%) - Levantine (24.1\%) - Gulf (18.3\%) - Sudan (13.0\%) - Iraqi (10.5\%) - Meghribi (6.4\%) & \\
         \cite{el-haj-2020-habibi} & \\
         \bottomrule
    \end{tabular}%
    \caption{The performance of regional-level ADI systems introduced in 8 different papers. The result of the best-performing model in each paper is reported. \textbf{Note}: \textsuperscript{\textbf{*}}: the exact number of samples in each split is not explicitly reported and the used data splits could not be found, \textsuperscript{$\dagger$}: the train/test sets are based on random sampling from the same dataset (i.e., the same data distribution), \textsuperscript{\textbf{Sp}}: the models' predictions are also based on additional speech features provided by the shared task organizers.}
    \label{tab:regional_ADI_results}
\end{table}

We contrasted the maximal estimated accuracy of 63.06\% to the results of 8 different regional-level ADI papers, summarized in \autoref{tab:regional_ADI_results}.
Two issues arise in analyzing the results, which might have led to inflated models' performances. First, 5 papers used random train/test splits. Consequently, the test set's samples come from the same distribution as the training set, which was previously found to be problematic \cite{sogaard-etal-2021-need}.
Second, five papers reported accuracy scores on imbalanced test sets, for which macro-averaged F1-scores are more appropriate.
Despite these two performance-inflating issues, all the reported scores still indicate that the task is not solved, except for the \textit{MADAR (Corpus-6)} dataset \cite{salameh-etal-2018-fine}, for which we identify two potential reasons. MADAR's authors identified Beirut, Cairo, Doha, Tunis, and Rabat as anchor cities for wider regional dialects. Hence, sentences written in these city-level dialects might have been more distinguishable from each other compared to sentences from other non-anchor cities. Moreover, the dataset was created by translating the same sentences from English or French into MSA in addition to the 5 city dialects. The translators might have tried to include more cues of their dialects in their translations to distinguish them from MSA translations and the other dialects' translations.

\section{Interannotator Agreement Scores for the Jordanian and Saudi Annotations}
\label{sec:IAA}

\begin{table}[h]
    \centering
    \tabcolsep5pt
    \small
    \begin{tabular}{lcccc}
    \multirow{2}{*}{\textbf{Country}} & \multicolumn{3}{c}{\textbf{Validity labels}} & \textbf{ALDi ratings}\\
    & \textbf{Fleiss} $\mathbf{\kappa}$ & \textbf{N valid} & \textbf{N ¬valid} & \textbf{Krip.} $\mathbf{\alpha}$ \\
    \midrule
        Jordan &     0.56 & 617 (455) &  503 (367) &     0.62 \\
        Saudi Arabia &     0.62 & 476 (328) &  644 (490) &     0.65 \\
    \bottomrule
    \end{tabular}
    \caption{The Interannotator agreement scores for the validity labels and ALDi ratings, Fleiss' Kappa ($\kappa$) for Validity labels and Krippendorff's Alpha --interval method-- ($\alpha$) for ALDi ratings. \textit{N valid} and \textit{N ¬valid} represent the number of samples whose majority vote labels are \textit{valid} and \textit{not valid}, respectively, with the number of sentences with complete agreement reported (between brackets).}
    \label{tab:IAA_agg}
\end{table}

We extended the annotations of the 1,120 samples of the NADI 2024 by recruiting 3 annotators from Jordan and 3 from Saudi Arabia. The interannotator agreement scores are reported in \autoref{tab:IAA_agg}. For the validity labels of each country, we compute the chance-corrected Fleiss' Kappa ($\kappa$) score, finding adequate agreement between the annotators of both countries. For the ALDi ratings, we use Krippendorff's Alpha --interval method-- ($\alpha$) between the numeric values of the ratings of each country's valid samples, which penalizes disagreements differently according to their assigned values. The range of the $\alpha$ scores is -1 to 1, with 0 indicating chance agreement. Hence, 0.62 and 0.65 signify that the annotators' agreement is substantially better than random, despite the subjectivity of the task.

\section{Country-level Overlap}
\label{sec:country_level_overlap}

\begin{figure}[h]
    \centering
    \includegraphics[]{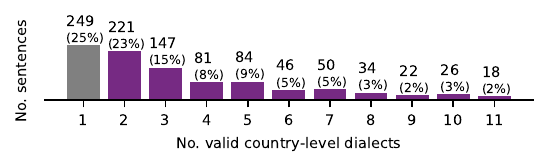}
    \caption{The histogram of the number of dialects in which a sentence is valid on the country-level dialects.}
    \label{fig:validity_countries}
\end{figure}

We compute the percentage of the samples within our dataset that are manually labeled as valid in multiple country-level dialects by annotators from these countries, to extend \citeposs{abdul-mageed-etal-2024-nadi} analysis by covering two additional country-level dialects.
Only 249 sentences ($\approx$ 25\%) are single-label as per \autoref{fig:validity_countries}, compared to the $\approx$ 30\% reported for 9 country-level dialects on NADI~2024's development set \citep{abdul-mageed-etal-2024-nadi}.  This indicated that incorporating more country-level dialects would still increase the already high percentage of multi-label samples.

\begin{figure}[h]
    \centering
    \includegraphics[width=.6\textwidth]{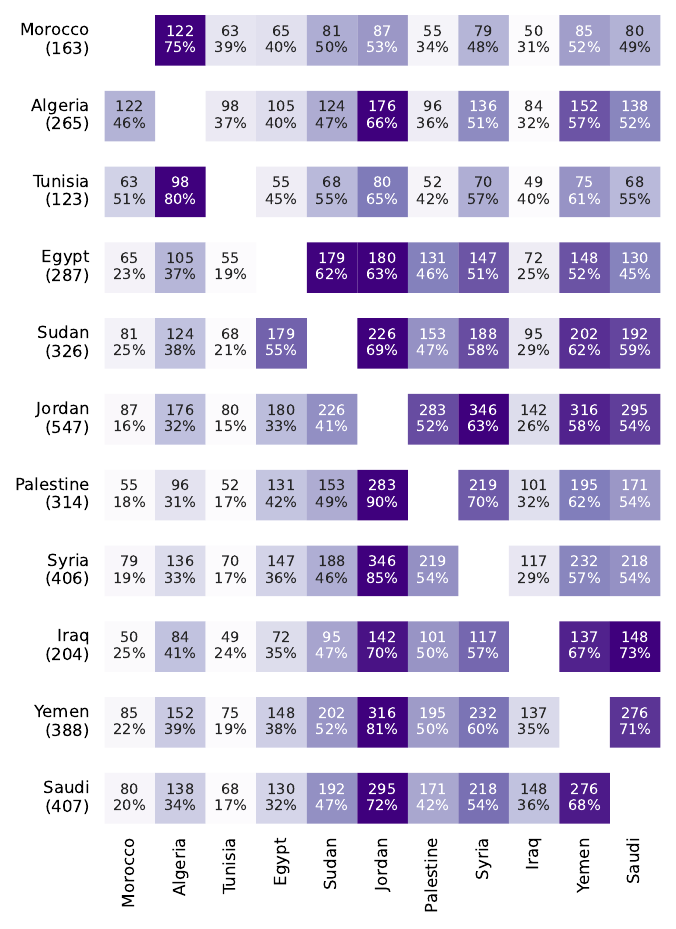}
    \caption{The percentage and number of each row country's valid samples that are also valid in the column country. \textbf{Note:} Each row's colormap range is independent from the other rows.}
    \label{fig:country_overlap}
\end{figure}

We also show the cross-country overlap in \autoref{fig:country_overlap}. While it is clear that countries within the same region overlap more with each other, a substantial overlap with countries from other regions exists. Theoretically, our dataset is uniformly representative of the 14 different countries to which the samples were geolocated. However, the NADI 2024's authors found that the precision of their geolocation methodology varies for the different countries, and is the lowest for the countries of the Maghreb region (49.3\% for Tunisia, 57.3\% for Morocco, and 65.3\% for Algeria). Hence, we think that further investigations are required before using these percentages as proxies for proximity between dialects.

\section{Lexical Cues}
\label{sec:lexical_cues_geolocation}

The TWT15DA is an ADI dataset built by iteratively augmenting lists of lexical cues of 15 country-level dialects using geolocated tweets having any of these cues, then streaming more geolocated tweets using the augmented lists \cite{althobaiti_2022}. For each country, the new cues to be added are non-MSA unigrams \textbf{(a)} in the tweets geolocated to this country, that \textbf{(b)} have high PMI values based on the following equation: $PMI(Unigram,\ Country) = log(\frac{P(Unigram,\ Country)}{P(Unigram) * P(Country)})$; where the probabilities are computed using maximum likelihood estimation. Therefore, the same unigram could have PMI scores for multiple countries (e.g., \ARA{كيفاش} /kyfAš/ in Algerian, Moroccan, and Tunisian Arabic lists with PMI scores of 2.07, 1.55, 1.19). Hence, these cues are not necessarily distinctive of a single country-level dialect. However, the author defines the \textit{cues} as ``words used in one or more Arabic dialects but never used in MSA, thereby distinguishing Arabic dialects from MSA''.

\begin{table}[tb]
        \centering
        \small    \begin{tabular}{c|cccc|ccc|cc}
         \textbf{Country} & \textbf{\samplesMatching} & \textbf{\samplesMatchingValid} & \textbf{\samplesMatchingExc} & \textbf{\samplesValid} & \textbf{P} &  \textbf{D} & \textbf{R} & \textbf{C} & \textbf{\cuesMatching} \\

        \midrule
Morocco &       52 &                  22 &                 5 &               163 & .42 & .10 & .13 &   410 &             45 \\
Algeria &       41 &                  23 &                 2 &               265 & .56 & .05 & .09 &   421 &             38 \\
Tunisia &       62 &                  19 &                 1 &               123 & .31 & .02 & .15 &   407 &             48 \\
  Egypt &       33 &                  23 &                18 &               287 & .70 & .55 & .08 &   172 &             35 \\
 Jordan &       51 &                  30 &                 1 &               547 & .59 & .02 & .05 &   180 &             37 \\
  Syria &       50 &                  28 &                 9 &               406 & .56 & .18 & .07 &    94 &             28 \\
   Iraq &       21 &                  13 &                12 &               204 & .62 & .57 & .06 &   179 &             18 \\
  Yemen &        8 &                   5 &                 2 &               388 & .62 & .25 & .01 &   137 &              8 \\
  Saudi &       43 &                  20 &                 6 &               407 & .47 & .14 & .05 &   145 &             26 \\
        \bottomrule
        \end{tabular}
        \caption{Lexical cues of the TWTDA15 datasets. \textbf{Note (1) :} For each region's list, we report the number of samples of our dataset matching any of the cues (\textit{\samplesMatching}) of which valid (\textit{\samplesMatchingValid})  and of which exclusively valid (\textit{\samplesMatchingExc}), in addition to the total number of valid samples (\textit{\samplesValid}). The last two columns represent the total number of regional cues (\textbf{C}) and the number of cues that match any of the samples (\textbf{\cuesMatching}). \textbf{Note (2):} The table lists the 9 countries that are common between the labels of our dataset, and the lists of TWT15DA which did not include \textit{Palestine} and \textit{Yemen}.}
        \label{tab:precision_TWT15DA_country}
\end{table}

We replicate the analysis in §\ref{sec:cues_analysis} for the TWT15DA dataset, and report the precision, recall, and distinctiveness scores in \autoref{tab:precision_TWT15DA_country}. Notably, the lists have a low range of precision scores $[0.31, 0.70]$, and an even lower range of distinctiveness scores $[0.02, 0.57]$.

\paragraph{Applying a Region's Lexical Cues only to the Region's Geolocated Samples} For the TWT15DA dataset, each sample should have at least a cue for one of the dialects. However, the assigned label is based on the sample's geolocation, and not on the dialects associated to the cues. Hence, to assign a sample to a country-level dialect, the sample should \textbf{(a)} have a lexical cue of this dialect and \textbf{(b)} be geolocated to this country.
To simulate this two-step method for each country's/region's list, we replicate our method, but then only consider the matching samples that are geolocated to the considered country/region.
The results of applying this post-processing step for the three lists of cues (DART, DIAL2MSA, and TWT15DA) are reported in \autoref{tab:precision_cues_geolocation}. The effectiveness of this step is better understood by contrasting the results in \autoref{tab:precision_cues} and \autoref{tab:precision_TWT15DA_country} to those in \autoref{tab:precision_cues_geolocation}.

\begin{table}[t!]
    \begin{subtable}[t]{0.45\columnwidth}
    \centering
    \small    \addtolength{\tabcolsep}{-2.5pt}    
    \begin{tabular}{c|cccc|ccc|cc}
        \textbf{Region} & \textbf{\samplesMatching} & \textbf{\samplesMatchingValid} & \textbf{\samplesMatchingExc} & \textbf{\samplesValid} & \textbf{P} &  \textbf{D} & \textbf{R} & \textbf{C} & \textbf{\cuesMatching} \\
        \midrule
   EGY &       20 &                  20 &                13 &               287 &        1.0 & .65 & .07 &   271 &             10 \\
   IRQ &        6 &                   6 &                 6 &               204 &        1.0 &          1.0 & .03 &   120 &              7 \\
         \midrule
   MGH &       15 &                  15 &                14 &               325 &       1.0 & .93 & .05 &   273 &             11 \\
   LEV &       24 &                  22 &                20 &               629 & .92 & .83 & .03 &   240 &              8 \\
   GLF &        0 &                   0 &                 0 &               407 &         - &            - & .00 &   200 &              0 \\
         \bottomrule
    \end{tabular}
        \label{tab:precision_DART_geolocation}
        \caption{DART's 5 regional lists.}
    \end{subtable}%
    \hspace{0.85cm}
    \begin{subtable}[t]{0.45\columnwidth}
    \centering
    \small    \addtolength{\tabcolsep}{-2.5pt}
    \begin{tabular}{c|cccc|ccc|cc}
        \textbf{Region} & \textbf{\samplesMatching} & \textbf{\samplesMatchingValid} & \textbf{\samplesMatchingExc} & \textbf{\samplesValid} & \textbf{P} &  \textbf{D} & \textbf{R} & \textbf{C} & \textbf{\cuesMatching} \\

         \midrule
   EGY &       25 &                  25 &                15 &               287 &        1.0 & .6 & .09 &    28 &             10 \\
   \midrule
   MGH &       34 &                  32 &                29 &               325 & .94 & .85 & .10 &    60 &             22 \\
   LEV &       36 &                  33 &                33 &               629 & .92 & .92 & .05 &    31 &             11 \\
   GLF &        0 &                   0 &                 0 &               407 &         - &            - & .00 &     9 &              0 \\
   \bottomrule
    \end{tabular}
        \label{tab:precision_DIAL2MSA_geolocation}
        \caption{DIAL2MSA's 4 regional lists.}
    \end{subtable}%
    
    \vspace{3mm}
    \centering
    \begin{subtable}[t]{0.45\columnwidth}
    \centering
    \small    \addtolength{\tabcolsep}{-2.5pt}
    \begin{tabular}{c|cccc|ccc|cc}
        \textbf{Country} & \textbf{\samplesMatching} & \textbf{\samplesMatchingValid} & \textbf{\samplesMatchingExc} & \textbf{\samplesValid} & \textbf{P} &  \textbf{D} & \textbf{R} & \textbf{C} & \textbf{\cuesMatching} \\
         \midrule
Morocco &       15 &                  14 &                 5 &               163 & .93 & .33 & .09 &   410 &             22 \\
Algeria &       13 &                  13 &                 2 &               265 &       1.0 & .15 & .05 &   421 &             18 \\
Tunisia &       12 &                   9 &                 1 &               123 & .75 & .08 & .07 &   407 &             15 \\
  Egypt &       13 &                  13 &                11 &               287 &       1.0 & .85 & .05 &   172 &             14 \\
 Jordan &       15 &                  12 &                 1 &               547 & .80 & .07 & .02 &   180 &             14 \\
  Syria &       12 &                  11 &                 5 &               406 & .92 & .42 & .03 &    94 &             12 \\
   Iraq &       11 &                  11 &                11 &               204 &       1.0 &          1.0 & .05 &   179 &             13 \\
  Yemen &        4 &                   4 &                 2 &               388 &       1.0 & .50 & .01 &   137 &              6 \\
  Saudi &        9 &                   9 &                 4 &               407 &       1.0 & .44 & .02 &   145 &              7 \\
		\bottomrule
    \end{tabular}
        \label{tab:precision_TWT15DA_geolocation}
        \caption{TWT15DA's 9 country-level lists.}
    \end{subtable}%
    \caption{The Precision (\textit{P}), Distinctiveness (\textit{D}), and Recall (\textit{R}) of each region's/country's cues, when the matching samples not geolocated to the region/country are discarded. \textbf{Note:} For each region's list, we report the number of samples geolocated to this region, matching any of its cues (\textit{\samplesMatching}) of which valid (\textit{\samplesMatchingValid})  and of which exclusively valid (\textit{\samplesMatchingExc}). The total number of samples valid in this regions (\textit{\samplesValid}) are reported irrespective of their geolocations. The last two columns represent the total number of regional cues (\textbf{C}) and the number of cues that match any of the samples (\textbf{\cuesMatching}).}
    \label{tab:precision_cues_geolocation}

\end{table}

The range of the precision significantly improves to values > 0.9 for the three lists, except for the lists of Tunisia and Jordan in TWT15DA. The distinctiveness scores also improve, yet to much lower ranges compared to the precision. This hints that filtering out the samples that match any of a region's cues, yet are not geolocated to this region minimizes the impact of matching false friends of these cues, which are intuitively expected to be in samples geolocated to other regions.

Unsuprisingly, limiting the samples to ones geolocated to each list's region causes a decrease in the recall values, as all the samples valid in this region's dialect that are not geolocated to the region are pre-filtered. Another drawback of this geolocation-based step is that the samples' geolocations are not always available.

\end{document}